%% file: main.tex
\documentclass[10pt,onecolumn,letterpaper]{article}

\usepackage{icb}
\usepackage{times}
\usepackage{epsfig}
\usepackage{graphicx}
\usepackage{amsmath}
\usepackage{amssymb}
\usepackage[hyphens]{url}   
\usepackage[colorlinks]{hyperref}

\icbfinalcopy 

\def\eq#1{(\ref{#1})}
\def\Vec#1{{\boldsymbol{#1}}}
\def\Mat#1{{\boldsymbol{#1}}}
\def\Df{{\operatorname{D}}}
\def\Gf{{\operatorname{G}}}
\def\log{\operatorname{log}}
\def\tilde{\raise.17ex\hbox{$\scriptstyle\mathtt{\sim}$}}

\ificbfinal\fi
\begin{document}

\title{TV-GAN: Generative Adversarial Network Based Thermal to Visible Face Recognition}

\author{Teng Zhang\footnotemark[1], Arnold Wiliem*, Siqi Yang and Brian C. Lovell\\
The University of Queensland, School of ITEE, QLD 4072, Australia\\
{\tt\small [patrick.zhang, a.williem, siqi.yang]@uq.edu.au, lovell@itee.uq.edu.au}
}

\footnotetext[1]{T. Zhang and A. Wiliem contributed equally to this work. Code will be available soon.}
\maketitle
\thispagestyle{empty}

\input{abstract}

\input{introduction}

\input{related_works}

\input{proposed_method}

\input{experiments}

\input{conclusion}

\input{ack}

{\small
\bibliographystyle{ieee}
\bibliography{egbib}
}

\end{document}

%% file: abstract.tex
\begin{abstract}
This work tackles the face recognition task on images captured using thermal camera sensors which can operate in the non-light environment.
While it can greatly increase the scope and benefits of the current security surveillance systems, performing such a task using thermal images is a challenging problem compared to face recognition task in the Visible Light Domain (VLD). 
This is partly due to the much smaller amount number of thermal imagery data collected compared to the VLD data. 
Unfortunately, direct application of the existing very strong face recognition models trained using VLD data 
into the thermal imagery data will not produce a satisfactory performance.
This is due to the existence of the domain gap between the thermal and VLD images.
To this end, we propose a Thermal-to-Visible Generative Adversarial Network (TV-GAN) that is able to transform thermal face images into their corresponding VLD images whilst maintaining identity information which is sufficient enough for the existing VLD face recognition models to perform recognition. Some examples are presented in Figure~\ref{figure_examples}.
Unlike the previous methods, our proposed TV-GAN uses an explicit closed-set face recognition loss to regularize the discriminator network training. 
This information will then be conveyed into the generator network in the forms of gradient loss.
In the experiment, we show that by using this additional explicit regularization for the discriminator network, the TV-GAN is able to preserve more identity information when translating a thermal image of a person which is not seen before by the TV-GAN.

\end{abstract}

%% file: introduction.tex
\section{Introduction}

\begin{figure}[htb]
  \centering
  \includegraphics[width=0.6 \linewidth]{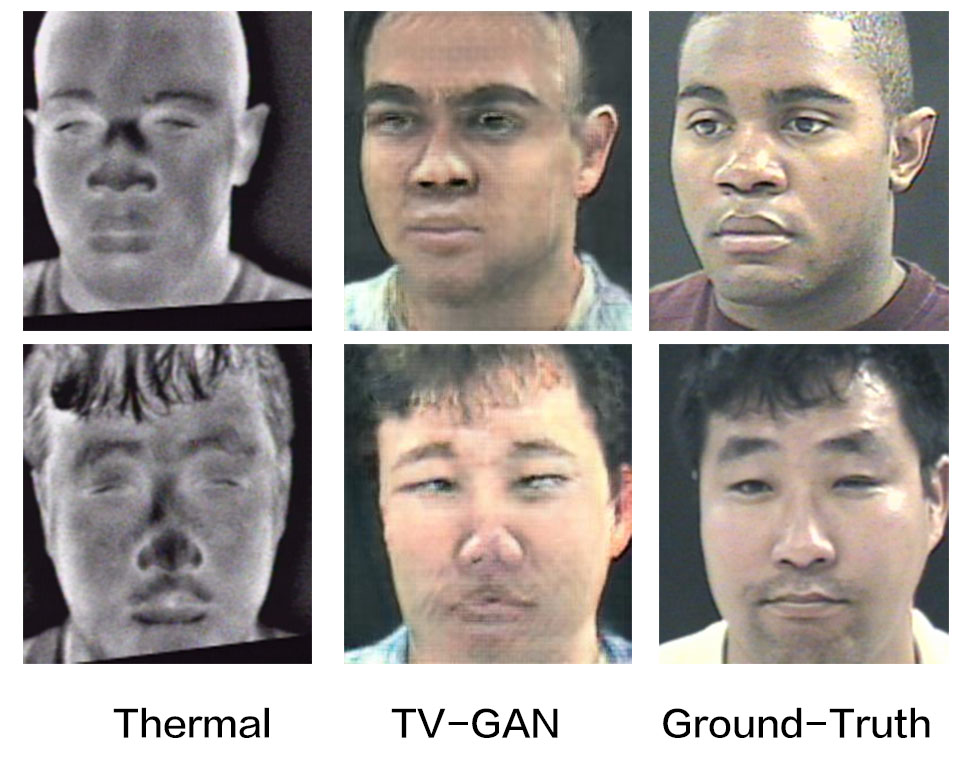}
  \caption{Example results. The first column is input thermal faces and the second column are the generated faces. Note that all these three people have never been seen in the model training phase. Comparing with the ground-truth from the last column, we can see that the proposed method can recover  considerable detail.}
  \label{figure_examples}
\end{figure}

Face recognition is one of the most important tasks in a smart video surveillance systems and it has been extensively studied in the visible light domain (VLD). Recently, the existing deep neural network based VLD face recognition systems have achieved impressive performance~\cite{Parkhi15, Schroff15, Wen2016}. This is due to the advent of extremely large face datasets~\cite{LFWTechUpdate, 2014Yi}. With these great strides, it is imperative to extend the existing VLD based face recognition systems into the other less studied domains such as near-infrared imaging (low-light) and thermal imaging (no-light).

\begin{figure}[htb]
  \centering
  \includegraphics[width=0.6 \linewidth]{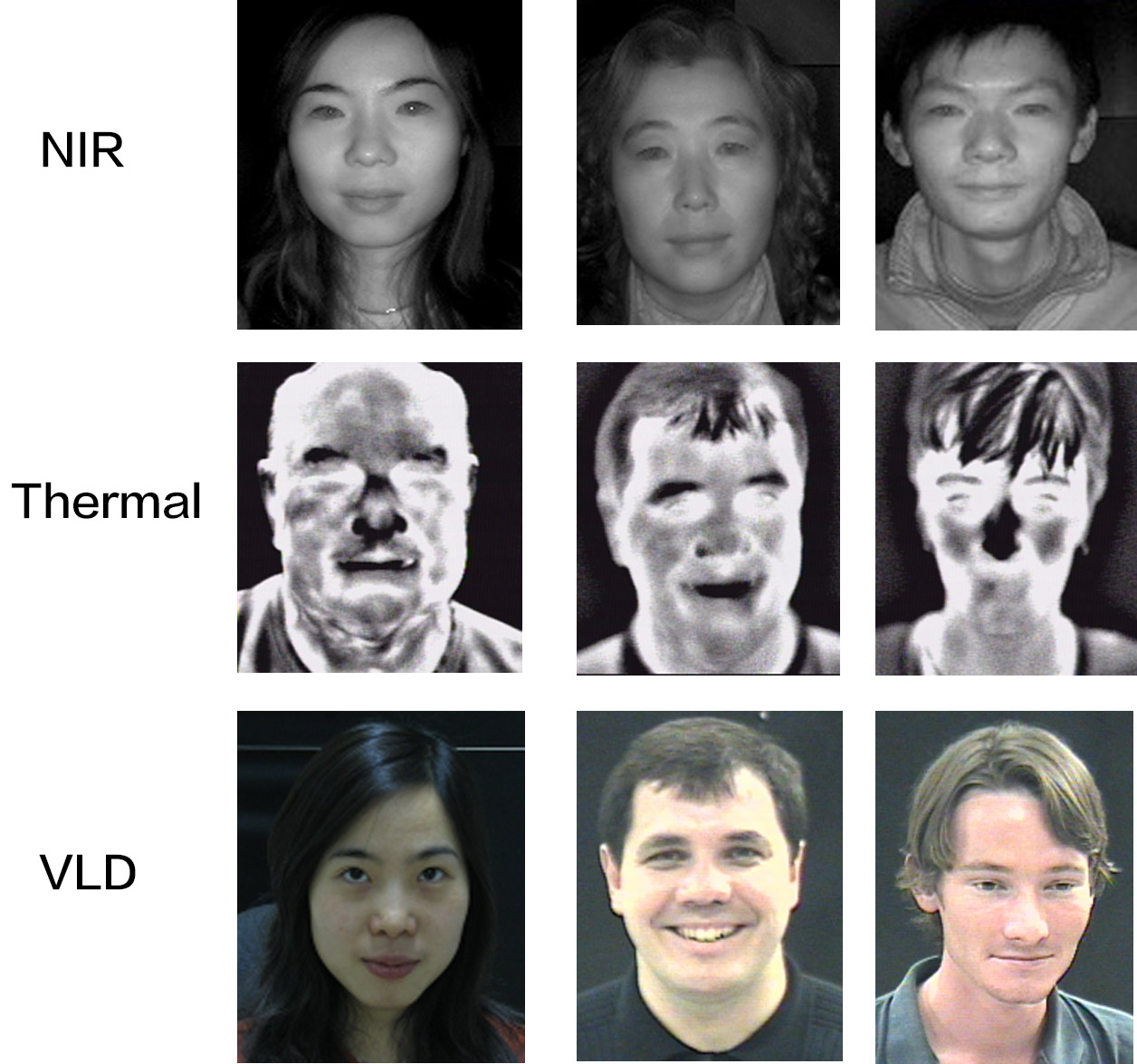}
  \caption{Randomly selected face image samples in three different domains.}
  \label{figure_diff_domains}
\end{figure}

The difference of these three domains is illustrated in Figure~\ref{figure_diff_domains}. 
Whilst, the thermal face has lost most of the texture and edge information, the near-infrared face has very much similarity to the VLD face images.
As such, performing face recognition task in the thermal image domain is significantly more challenging than in the near-infrared image domain.

Unfortunately, the above-mentioned successes in the VLD domain could not be easily replicated in the thermal domain due to relatively small amount of training data available in this domain and the domain gap between the thermal and the visible light. As shown in our experiment, directly applying the VLD based face recognition systems will not achieve satisfactory performance.

\begin{figure}[htb]
  \centering
  \includegraphics[width=0.8 \linewidth]{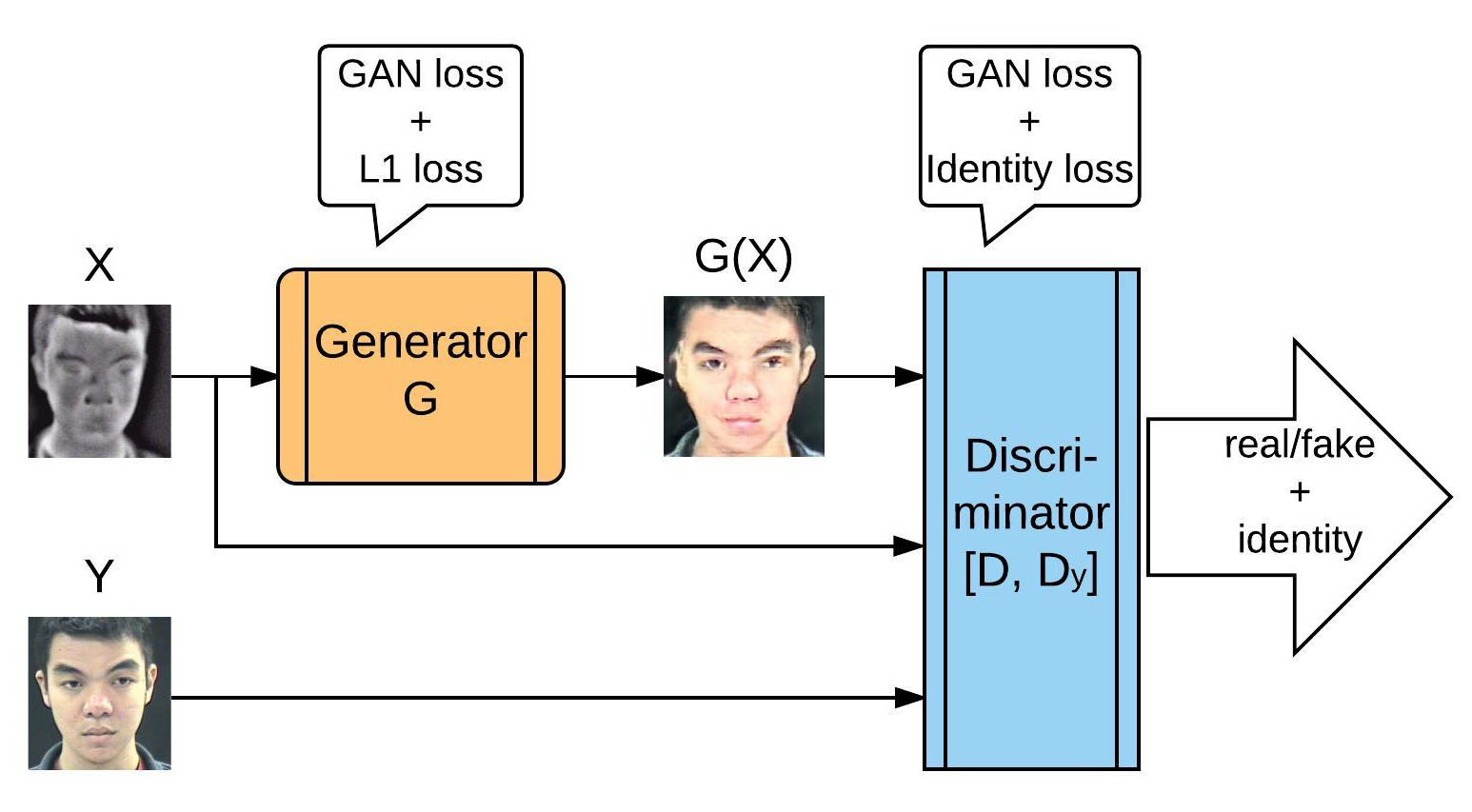}
  \caption{The proposed TV-GAN structure: Our discriminator not only provides the discrimination of fake and real, but also performs a closed-set face recognition task. The generator network $\Gf$ aims to produce images (called ``fake'' images) that can ``fool'' the discriminator. Note that our multi-task discriminator is composed of two networks with shared weights $[ \Df, \Df_{\Vec{y}} ]$, where $\Df$ discriminates whether $\Mat{Y}$ is real or not and $\Df_{\Vec{y}}: \Mat{X} \times \Mat{Y} \mapsto \{0,1\}^{N+1}$ performs the closed-set face recognition task.}
  \label{figure_framework}
 
\end{figure}

To this end, our strategy is to utilize image transformation techniques to the thermal query images.
Once transformed, an existing pre-trained VLD deep neural network face recognition model can be directly employed as a black box. 
So, instead of using the thermal face, we use the hallucinated VLD face fed into the VLD face recognition model.
With this pre-processing strategy, we can achieve much better recognition results without changing or retraining the VLD model. The framework of our proposed method is sketched in Figure~\ref{figure_framework}. 

To achieve good recognition performance, we train a generator network using the Generative Adversarial Network (GAN). 
Here, we use an adversary network, here denoted as a discriminator network aiming to discriminate between real or fake/generated images to train the generator.
In order to preserve the identity information of the person in the transformed image, we add an identity loss which is based on the closed-set face recognition task into the discriminator network. 
Unlike the previous work in~\cite{ZhangIJCB2017} that uses perception loss~\cite{Johnson2017} to indirectly measure the identity loss, our identity loss is more explicit.
Note that the perception loss requires training another classifier network such as VGG network~\cite{Simonyan2015} and thus makes the training significantly more expensive.

\textbf{Contributions - } Our contributions can be listed as follows.
The main contribution is that we propose a novel Thermal-to-Visible Generative Adversarial Network (TV-GAN) that is able to preserve sufficient identity information when transforming thermal images into their corresponding VLD images. We draw our inspiration from the recent Pix2Pix~\cite{Isola2016} and DR-GAN~\cite{Tran2017}.
Our key insight is to train our discriminator to perform both the binary classification of fake or real and to perform a closed-set face recognition task. 
We validate our method in challenging experiment scenarios and still achieve good performance comparing to the other recent methods.

This paper is continued as follows: Section 2 will introduce the related works and the proposed approach is described in Section 3. Experimental results are provided and discussed in Section 4. Finally, we conclude this paper in Section 5.

%% file: related_works.tex
\section{Related Works}

As discussed above, face recognition is an extensively studied area in the visible light domain. Recently, many researchers have attempted to address the same problem in other light domains such as near-infrared domain~\cite{Yi2007}~\cite{Juefei15}~\cite{Lezama16}, polar-metric thermal domain~\cite{Hu2016}~\cite{Riggan2016}~\cite{Nathaniel15} and conventional thermal domain~\cite{Sarfraz2017}~\cite{Reale2014}~\cite{Mostafa13}.

Before the deep learning era, a few works primarily aiming at training traditional classifiers and feature representation were proposed. In~\cite{Li2008}, a simple method that hallucinates the visible face from the thermal face was proposed by exploiting the local linearity in not only the image spatial domain but also the image manifolds. 
Choi~\etal~\cite{Jonghyun12} presented a partial least square (PLS) based regression framework to model the large modality difference in the PLS latent space. 
Similarly, Hu~\etal~\cite{Shuowen15} used a discriminant PLS based approach by specifically building discriminant PLS gallery models for each subject by using thermal cross examples.

Recent methods utilizing deep neural networks have been proposed. For instance, Sarfraz~\etal~\cite{Sarfraz2017} used a neural network to learn the reverse mapping, from VLD to mid-wave and low-wave infrared, so that a thermal face image could be matched to its VLD counterpart. 
This strategy has the disadvantage of having to apply the mapping to each VLD image in the dataset. 
We propose to use a conditional generative adversarial network to transform a query thermal image into the VLD image. 

A possible approach to address this is by reducing the domain gap between the thermal and visible light domains.
This can be done by training a deep neural network that can transform thermal images into the corresponding visible images.
For instance, Lezama~\etal~\cite{Lezama16} proposed to use a patch-based transform CNN to hallucinate a visible face from a near-infrared face. 
The work from~\cite{Isola2016}, known as Pix2Pix, can transform images from domain A to domain B. The idea is to train a Conditional Generative Adversarial Network (CGAN) that can ``fool" domain B classifier using a processed image from domain A. However, this work did not ensure that the identity information was preserved during the transformation.

Perhaps the most similar work to us is the work from Zhang~\etal ~\cite{ZhangIJCB2017} which also employed GAN to generate faces from the thermal input. However, there are significant differences between our framework and~\cite{ZhangIJCB2017}. Firstly, their work calculates the identity loss indirectly using high-level semantic features extracted from a classification network. This requires additional network during training. Secondly, we consider a much more challenging scenario wherein the visible light images are in color instead of grey scale and the faces have various pose and occlusion with eyeglasses. In addition, the images used in our evaluation are not perfectly aligned and included head, neck, and part of the chest.

%% file: proposed_method.tex
\section{Proposed Framework}

In this section, a brief description of the Generative Adversarial Networks will be presented. Then, the proposed TV-GAN is elucidated.

\subsection{Generative Adversarial Networks}

Generative Adversarial Network (GAN) is first introduced in~\cite{Goodfellow2014}. It comprises two models with competing tasks. The first model, the generator model $\Gf$, aims to generate an image which resemblances a real image. The aim for the second model, the discriminator $\Df$ is to separate between the fake images from the generator and the real images. Generally, both models are represented by deep neural networks. 

Since its first introduction, GAN has been extended into various applications. For instance, Mirza~\etal~\cite{Mirza2014} propose the Conditional GAN of which the generator learns the data distribution condition upon an input. Radford~\etal~\cite{Radford2016} proposed a class of GAN that can stabilize the training. Recently GAN has also been extended for generating images from text description~\cite{Zhang2017}, generating style and structure of natural indoor scene images~\cite{Wang2016} and translating an image from one domain to the other~\cite{Isola2016}.

In this work, we use the conditional GAN framework that allows transforming an image from one domain to the other domain.
The GAN architecture used in this work can be briefly described as follows. Let $\Df : \Mat{X} \times \Mat{Y} \mapsto \{ fake, real\}$, $\Mat{X}, \Mat{Y} \in \mathbb{R}^{w \times h}$ be the discriminator function and $\Gf: \Mat{X} \mapsto \Mat{Y}$ be the generator. Note that, unlike the original GAN description in~\cite{Goodfellow2014} using a Gaussian random vector $\Vec{z}$ as the input for the generator function, we follow the GAN architecture described in~\cite{Isola2016} which uses dropout to maintain the sample diversity. 
In addition, according to~\cite{Isola2016}, the generator $\Gf$ will generate a better image when it is trained with a discriminator $\Df$ admitting two inputs: the original image $\Mat{X}$, and the transformed image $\Mat{Y}$. The transformed image $\Mat{Y}$ is from the ground truth when $\Mat{Y} \tilde P_{data}$, where $P_{data}$ is the distribution generated from a set of real images, or generated by using $\Gf ( \Mat{X} )$.
The architecture is then trained using the following objective:
\begin{align}
    \mathcal{L}_{cGAN}(\Gf,\Df) = \mathbb{E}_{\Mat{X} \tilde P_{data}(\Mat{X})} \left[ \log  \Df ( \Mat{X} , \Mat{Y} ) \right] + \nonumber \\
    \mathbb{E}_{\Mat{X} \tilde  P_{data}} \left[ \log  1 - \Df ( \Mat{X} , \Gf ( \Mat{X} ) \right]
    \label{eq:cGAN}
\end{align}
\noindent
where $\mathcal{L}_{cGAN}$ is the conditional GAN loss function.

\subsection{TV-GAN: Thermal-to-Visible GAN}

The goal of Thermal-to-Visible GAN (TV-GAN) is to train a generator $\Gf$ that will transform a thermal image $\Mat{X}$ into its corresponding visible image $\Mat{Y}$ of which the visible image $\Mat{Y}$ still carries sufficient identity information for face recognition task. 
To this end, we base our method on Pix2Pix~\cite{Isola2016}. Pix2Pix is able to transform an image from one domain to the other domain. 
Unlike the CGAN, Pix2Pix is able to generate sharper images due to its additional loss function that explicitly penalizes 
the deviation of the generated image $\Gf ( \Mat{X} )$ from the ground truth $\Mat{Y} \tilde P_{data}$.
Unfortunately, Pix2Pix does not have explicit regularization that helps to preserve the personal identity. 
Recent work in~\cite{Tran2017} that proposes Disentangled-Representation GAN (DR-GAN) shows that it is possible to improve feature discrimination for face recognition task by explicitly adding identity loss function to the discriminator training loss function. 
The efficacy of using identity loss has also been shown in the GAN-based Visible Face Synthesis (GAN-VFS)~\cite{ZhangIJCB2017}. The difference is that GAN-VFS calculates the loss indirectly by using the perceptual loss~\cite{Johnson2017} which uses high-level semantic features extracted from a classification network such as the VGG network~\cite{Simonyan2015}.

Different from GAN-VFS, we use the more explicit identity loss function similar to the DR-GAN which is aimed to learn disentangled feature representation solely from VLD images.
More specifically, we define our discriminator as a multi-task discriminator that does not only provide the discrimination fake or real but also performs a closed-set face recognition task. We note that although we train the discriminator to perform a closed-set face recognition task, the aim here is to use the gradient information from the discriminator to train the generator so it can generate visible images with sufficient identity information of the person for the recognition task.
Later in the experiment part we will show that this approach is still effective for performing the face recognition tasks where the query person has not been seen by the TV-GAN.
Let $\Vec{y} \in \{0,1\}^{N+1}$ be a one-hot-encoding $(N+1)$-dimensional identity vector wherein if the $p$-th element of vector $\Vec{y}_i$ is 1, then the image $\Mat{X}_i$ belongs to the $p$-th person; $N$ is the number of subjects in the training set and we reserve additional dimension for the generated images. This way, the discriminator only learns the identity information from the real images.
Our multi-task discriminator is composed of two networks with shared weights $[ \Df, \Df_{\Vec{y}} ]$, where $\Df$ discriminates whether $\Mat{Y}$ is real or not and $\Df_{\Vec{y}}: \Mat{X} \times \Mat{Y} \mapsto \{0,1\}^{N+1}$ performs the closed-set face recognition task.

The proposed TV-GAN training loss is defined as follows:
\begin{align}
& \mathcal{L}_{TV-GAN}(\Gf, \Df, \Df_{\Vec{y}}) = \nonumber \\ 
& \mathcal{L}_{cGAN}(\Gf, \Df) + \lambda_1 \mathcal{L}_{\ell_1}(\Gf) + \lambda_2 \mathcal{L}_{id} (\Gf, \Df_{\Vec{y}}) , 
\end{align}
\noindent
where $\mathcal{L}_{cGAN}$ is defined in~\eq{eq:cGAN}, $\mathcal{L}_{\ell_1}$ is the additional loss function from Pix2Pix defined as:
\begin{equation}
\mathcal{L}_{\ell_1}(\Gf) =  \mathbb{E}_{\Mat{X}, \Mat{Y} \tilde P_{data}} \left[ \| \Mat{Y} - \Df ( \Mat{X} ) \|_1 \right] . 
\end{equation}
\noindent
The identity loss function is defined as follows:
\begin{equation}
\mathcal{L}_{id} (\Gf, \Df_{\Vec{y}}) = \mathbb{E}_{\Mat{X}, \Mat{Y} \tilde P_{data}} \left[ \log (\Df_{\Vec{y}} ( \Mat{X}, \Mat{Y} )  \right]
\end{equation}

As for the network architecture, we adopt both the Pix2Pix's generator and discriminator networks without modification. In particular, the generator network uses the U-Net network~\cite{Ronneberger2015}, which is an encoder-decoder with skip connection between mirrored layers in the encoder and decoder stacks.

%% file: experiments.tex
\section{Experiments and Results}

In this section, we first describe the implementation details and baselines. Then, the dataset and evaluation protocol will be presented. Finally, we provide analysis based on the performance of various methods. 

\subsection{Implementation}

All evaluations were done by using NVidia K40c GPU with the tensorflow framework. 
In addition, Adam optimizer~\cite{Adam2014} was used with a batch size of 1. 
Following~\cite{Isola2016}, all networks were trained from the scratch with learning rate of $0.0002$, $\beta_1 = 0.5$. As for TV-GAN, we trained the network with 65 epochs. The hyperparameters $\lambda_1$ and $\lambda_2$ in~\cite{Goodfellow2014} were set to 100 to make the loss terms in the same scale.

As for the VLD face recognition network, we used pre-trained MatConvNet VGG-based model from~\cite{Parkhi15} without any fine-tuning. We call this VGG-face.
The query of the VGG-face is a transformed image $\Mat{Y} = f ( \Mat{X} )$, where $\Mat{X}$ is the image in the thermal domain and $\Mat{Y}$ is the transformed image. All the images in the gallery $\mathcal{G} = \{ \Mat{X}_m \}_{m=1}^M$ are VLD images.

\subsection{Baselines}

In the evaluation, three baselines were used:

\noindent
\textbf{Plain Thermal - } No transformation was applied on this baseline. In other words, for this baseline, the function $f$ only does the identity mapping, $\Mat{Y} = \Mat{X}$. Thus, essentially, this baseline will indicate the effect of the domain gap between thermal and visible light to the face recognition models such as VGG-face trained solely under VLD images.

\noindent
\textbf{Patch based method - } It has been shown in~\cite{Lezama16} that it is possible to learn transformation function $f ( \cdot )$ for Near Infrared Domain to VLD by using CNN based with encoder-decoder architecture. We apply this for thermal-to-visible conversion. More specifically, a set of paired image patches were first extracted and then the CNN was trained based on these patches. In this experiment, the patch size $25 \times 25$ was used. For a fair comparison, we did not apply the post-processing method blending images from both domains to obtain better performance. As for the CNN architecture, we opted to use a more recent CNN architecture called RedNet~\cite{RedNet16} which shares similarities to the U-Net~\cite{Ronneberger2015}. The difference is that RedNet has a skip connection the same as ResNet~\cite{resNet16} whereas U-Net has a skip connection the same as DenseNet~\cite{denseNet17}. From our empirical evaluation (not shown here) both skip connection types gave similar performance. We used RedNet20 which has 20 layers and trained with 108 epochs. The difference between the Patch-based Transform and TV-GAN generator function $\Gf$ is how the networks are trained. The Patch-based Transform method used the mean squared error loss, whereas TV-GAN used Generative Adversarial loss. 

\noindent
\textbf{Pix2Pix~\cite{Isola2016} - } As mentioned in the previous section that Pix2Pix does not have the identity loss regularization in the training whilst the proposed TV-GAN has this regularization. Since Pix2Pix is a GAN-based method, the transformation function $f$ is its generator function, $f = \Gf$. We trained Pix2Pix using 85 epochs.

\subsection{Dataset and evaluation protocol}

We used the IRIS dataset~\cite{irisdataset} for the evaluation in which all images were captured by FLIR SC6700 (spectral range 3um -- 7 um).
There were 29 subjects with 4,228 pairs of thermal/visible images.
As the subjects have various poses, we excluded repeated angles, extreme poses, expressions, and illumination for our experiments.
In total, there were 695 pairs of roughly aligned thermal/visible images ($695 \times 2 = 1,390$ images in total) of 29 subjects used in our experiments for all methods.

In the experiment, we needed to measure how much identity information preserved by the transformation function $f$ for each method. 
To this end, recognition accuracy metric was used. 
We measured this based on the accuracy of the VGG-face in recognizing the query $\Mat{Y} = f ( \Mat{X} )$ with the gallery $\mathcal{G}$.
More specifically, we used VGG-face to extract features from $\Mat{Y}$.
Then, nearest neighbor search with cosine distance metric was used.

There were two protocols used in the evaluation.
Protocol A: each subject in the gallery only had one frontal face; and Protocol B: each person in the gallery had four images covering several pose angles. The experiment was repeated multiple times and the average performance was reported. In addition, we also reported rank one, three, five and seven performance.
For each repeat, the dataset was randomly divided into 8 subjects for testing and 21 subjects for training the transformation function $f$ (approximately 500 images). 
This protocol ensured that no subject images were in training and testing sets. 
The gallery used by the VGG-face always contained 29 subjects with a different number of images (\ie one image each subject for protocol A and four images each subject for protocol B).
The test set for each split was used as the queries for the VGG-face. TV-GAN and Pix2Pix were trained on each split training set.
For a fair comparison, we did not perform any data augmentation such as horizontal flipping, rotation and cropping for the gallery images used by the VGG-face. The data augmentation was only applied for training the transformation function $f$.

\subsection{Results}

\begin{table}[htbp]
\centering
\caption{Average recognition accuracy (in \%) for the setting where only one frontal face visible image for each person is available (Protocol A).}
\begin{tabular}{l c c c c }
\hline
Accuracy  &Rank 1 &Rank 3 &Rank 5 &Rank 7\\ \hline
Plain Thermal      &3.5  &12.2  &21.9  &27.2\\ \hline
Patch based        &8.9  &18.9 &26.9 &36\\ \hline
Pix2Pix            &12.1 &28.8 &39.2 &47 \\ \hline
TV-GAN  &\textbf{13.9} &\textbf{33} &\textbf{46.8}  &\textbf{53.4}\\ \hline
\end{tabular}

\label{table_setting_a}
\end{table}

\begin{table}[htbp]
\centering
\caption{Average recognition accuracy (in \%) for the setting where four face visible images for each person are available (Protocol B)}
\begin{tabular}{l c c c c }
\hline
Accuracy  &Rank 1 &Rank 3 &Rank 5 &Rank 7\\ \hline
Plain Thermal  &4.9 &15.6  &21.45  &26.5\\ \hline
Patch based    &14.6  &23.5 &30.1 &35.7 \\ \hline
Pix2Pix        &16 &30.7  &37.3  &44.9 \\ \hline
TV-GAN   &\textbf{19.9} &\textbf{35.8} &\textbf{45.6} & \textbf{50.9}\\ \hline
\end{tabular}
\label{table_setting_b}
\end{table}

\begin{figure}[htb]
  \centering
  \includegraphics[width=0.6 \linewidth]{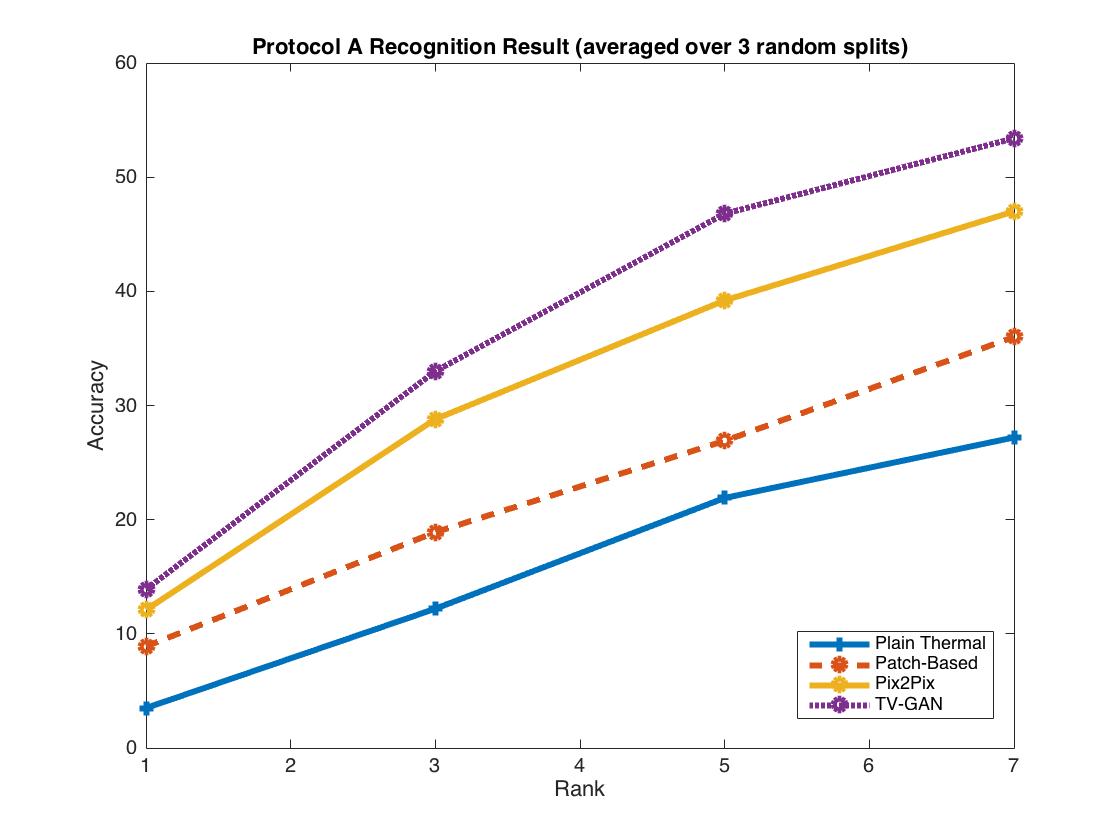}
  \caption{The gallery setting: one visible light image per person (protocol A). We run 3 splits and calculate the average results.}
  \label{figure_setting_a}
\end{figure}

\begin{figure}[htb]
  \centering
  \includegraphics[width=0.6 \linewidth]{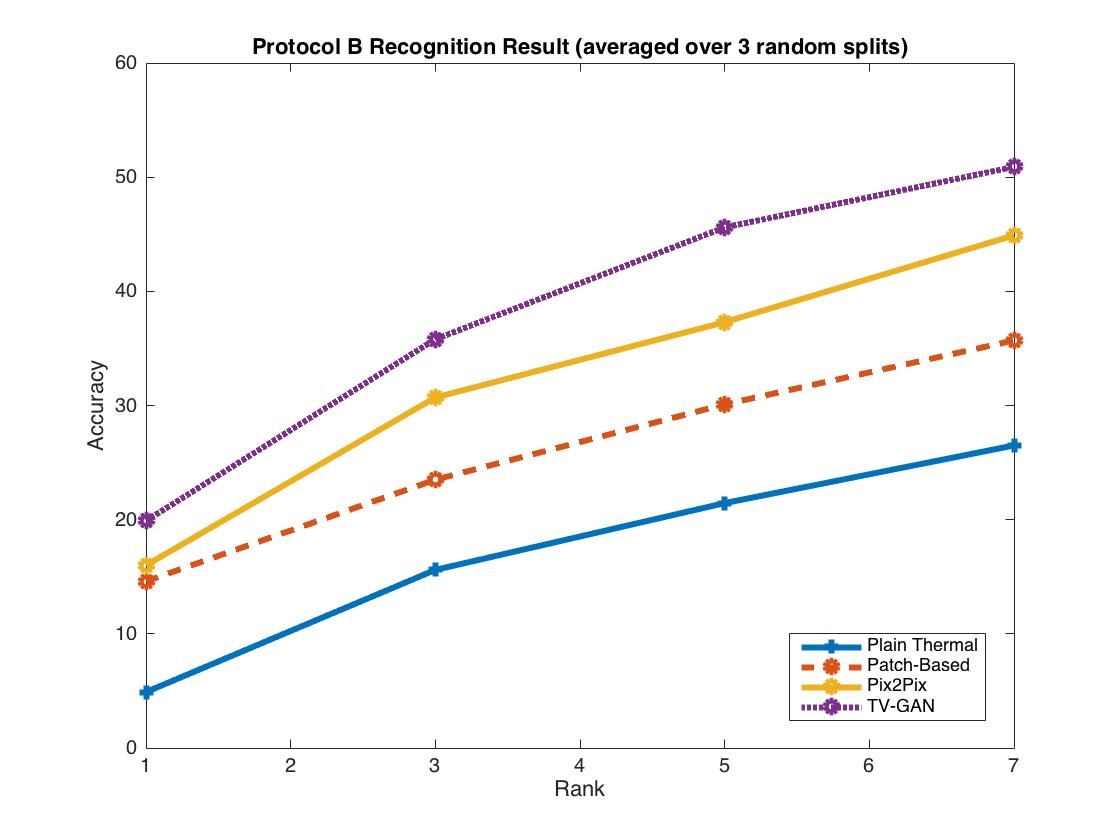}
  \caption{The gallery setting: four visible light image per person (protocol B). We run 3 splits and calculate the average results.}
 \label{figure_setting_b}
\end{figure}

The results for protocol A are presented in Table~\ref{table_setting_a} and Figure ~\ref{figure_setting_a}. The results for protocol B are reported in Table~\ref{table_setting_b} and Figure~\ref{figure_setting_b}. 
As we can see from these results the proposed TV-GAN outperforms all the methods. Also, it is noteworthy to mention that all methods outperform the plain thermal method by a large margin. This suggests the importance of reducing the domain gap existed between thermal and VLD images if one plans to utilize face recognition system trained solely in the VLD for recognizing faces in the thermal domain. 

The improvement from patch-based method to GAN-based methods such as Pix2pix and the proposed TV-GAN suggests the efficacy of the GAN loss for this application. Upon a closer look, the generators trained using GAN loss could generate much better images compared to the patch-based method. We conjecture that this might be caused by the following factors: (1) The thermal images may not carry sufficient information compared to the near infrared images; (2) the dataset contains large pose angles and (3) the paired thermal and visible images in the training data are not well aligned. As stated in~\cite{Lezama16} that this method requires well-aligned pair of images.

Finally, as the proposed TV-GAN outperforms Pix2Pix, this indicates that the closed-set face recognition regularization is more effective to train the generator that can preserve the personal identity in the generated images.

\begin{figure*}[htb]
  \centering\includegraphics[width=1 \linewidth]{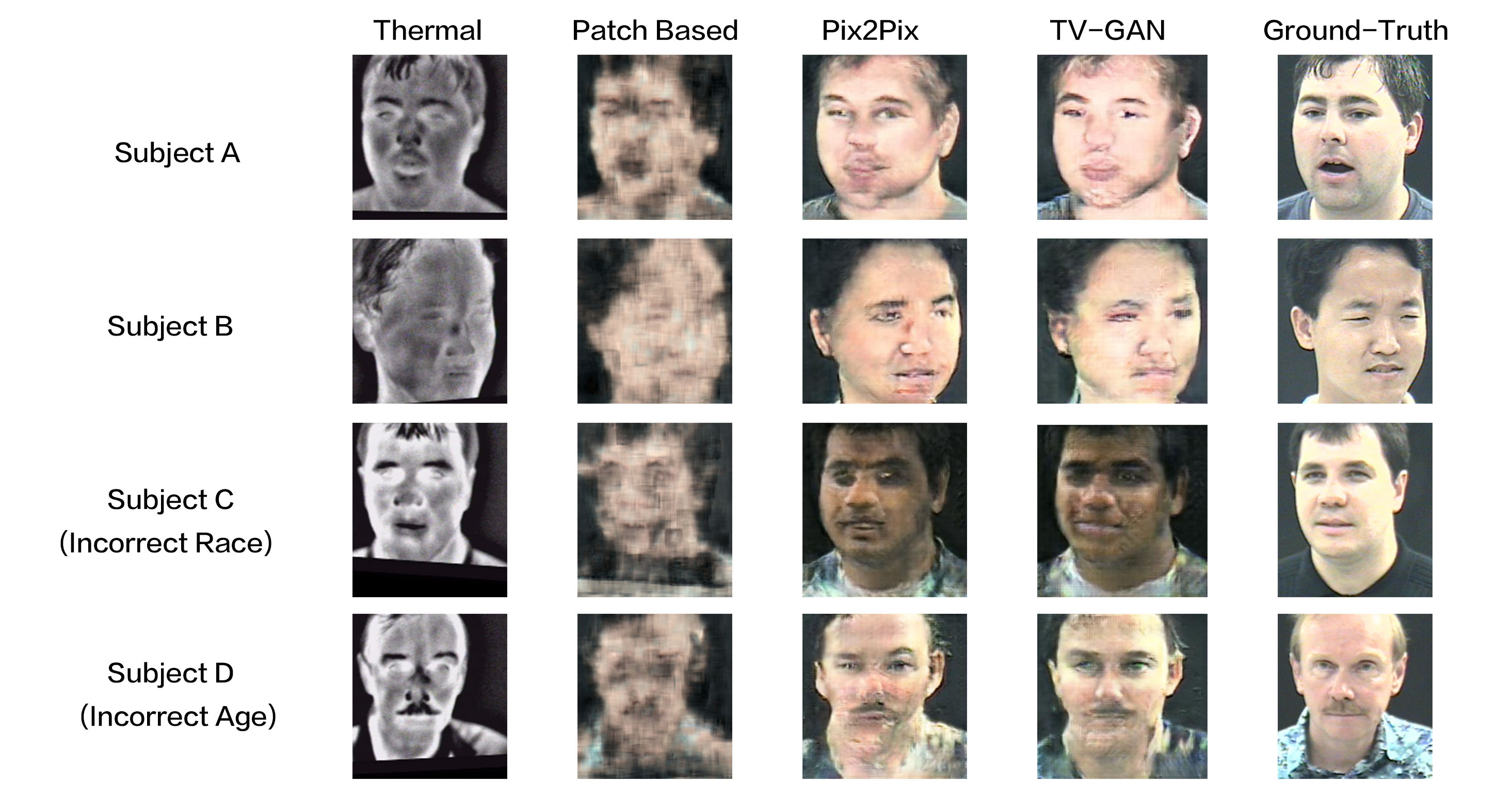}\par
  \caption{Visualization of selected results}
  \label{figure_visualization}
\end{figure*}%

\textbf{Visualization - } Here we present some visualization of all the methods. As we can see from Figure~\ref{figure_visualization}, the patch-based method generate the worst visible images. Both Pix2Pix and TV-GAN are able to generate reasonable visualization. However, it is noteworthy to mention that both TV-GAN and Pix2Pix do not always generate the correct attributes. For instance, the race of subject C and the age of subject D are incorrectly inferred. This could be due to the absence of regularization guiding the GAN training in preserving these face attributes. 

\begin{small}
\begin{table}[htbp]
\centering
\caption{We deliberately exclude all people with eye glasses from training set but put them all in the testing set to undermine the generator network of GAN.}
\begin{tabular}{l c c c c }
\hline
Accuracy  &Rank 1 &Rank 3 &Rank 5 &Rank 7\\ \hline
Plain Thermal  &\textbf{12.57}  &\textbf{36.65}  &\textbf{46.07}   &\textbf{54.97}\\ \hline
Pix2Pix        &7.85  &22.51  &36.13   &41.88 \\ \hline
TV-GAN   &12.04  &24.08  &31.41  &37.7\\ \hline
\end{tabular}
\label{table_no_glasses}
\end{table}
\end{small}

\textbf{Role of training data - } Despite its good performance, the proposed TV-GAN method is trained using a relatively small amount of images.  
Here we show the importance of having sufficient amount training data. To emulate data starvation environment we deliberately exclude all people with eyeglasses from the training data. The results are reported in Table~\ref{table_no_glasses}. The results suggest that both Pix2Pix and TV-GAN could not achieve good performance. Surprisingly, the plain thermal method outperforms both GAN methods. This could be due to the fact that the subjects in the testing set wear eyeglasses in both query and gallery images. This information may have been picked up by the features extracted from the VGG-face.

%% file: conclusion.tex
\section{Conclusion}

Thermal to visible face recognition/verification task is a challenging problem as not many data available and applying face recognition model exclusively trained on visible light domain images produces a non-satisfactory performance.
In this paper, we developed a GAN-based method that can apply pre-trained VLD deep learning models with no further fine tuning. 
More specifically, a generator network was trained using the Generative Adversarial Network framework which has two networks such as generator and discriminator networks trained against each other. 
The discriminator guided the generator via its gradient information.
Our key insight was that by using a closed-set face recognition task loss inserted into the discriminator, 
it allowed the generator learned the transformation function that preserved sufficient identity information for the VLD face recognition system.
Despite the existence of challenges such as occlusions, high pose, different skin tone and limited training data, in our experiment, we showed that our TV-GAN method outperformed the other methods.

Our proposed TV-GAN method is still far from perfect as it still did not ensure the correct transfer of the other face attributes such as race and age in the transformed images. This will be investigated in the future.

%% file: ack.tex
\subsection{Acknowledgement}

This work has been funded by the Australian Research Council (ARC) Linkage Projects Grant LP160101797. Arnold Wiliem is funded by the Advance Queensland Early-Career Research Fellowship. 

%% file: main.bbl
\begin{thebibliography}{10}\itemsep=-1pt

\bibitem{Jonghyun12}
J.~Choi, S.~Hu, S.~S. Young, and L.~S. Davis.
\newblock Thermal to visible face recognition.
\newblock volume 8371, pages 8371 -- 8371 -- 10, 2012.

\bibitem{Goodfellow2014}
I.~Goodfellow, J.~Pouget-Abadie, M.~Mirza, B.~Xu, D.~Warde-Farley, S.~Ozair,
  A.~Courville, and Y.~Bengio.
\newblock Generative adversarial networks.
\newblock In {\em NIPS}, pages 2672--2680, 2014.

\bibitem{resNet16}
K.~He, X.~Zhang, S.~Ren, and J.~Sun.
\newblock Deep residual learning for image recognition.
\newblock In {\em CVPR}, 2016.

\bibitem{Shuowen15}
S.~Hu, J.~Choi, A.~L. Chan, and W.~R. Schwartz.
\newblock Thermal-to-visible face recognition using partial least squares.
\newblock {\em Journal of the Optical Society of America A}, 32(3):431--442,
  2015.

\bibitem{Hu2016}
S.~Hu, N.~J. Short, B.~S. Riggan, C.~Gordon, K.~P. Gurton, M.~Thielke,
  P.~Gurram, and A.~L. Chan.
\newblock A polarimetric thermal database for face recognition research.
\newblock In {\em CVPRW}, pages 187--194, 2016.

\bibitem{denseNet17}
G.~Huang, Z.~Liu, L.~van~der Maaten, and K.~Q. Weinberger.
\newblock Densely connected convolutional networks.
\newblock In {\em CVPR}, 2017.

\bibitem{Isola2016}
P.~{Isola}, J.-Y. {Zhu}, T.~{Zhou}, and A.~A. {Efros}.
\newblock Image-to-image translation with conditional adversarial networks.
\newblock In {\em CVPR}, 2017.

\bibitem{Johnson2017}
J.~Johnson, A.~Alahi, and L.~Fei-Fei.
\newblock Perceptual losses for real-time style transfer and super-resolution.
\newblock In {\em ECCV}, 2016.

\bibitem{Juefei15}
F.~Juefei-Xu, D.~K. Pal, and M.~Savvides.
\newblock Nir-vis heterogeneous face recognition via cross-spectral joint
  dictionary learning and reconstruction.
\newblock In {\em CVPRW}, pages 141--150, 2015.

\bibitem{Adam2014}
D.~P. Kingma and J.~Ba.
\newblock Adam: A method for stochastic optimization.
\newblock In {\em ICLR}, 2014.

\bibitem{LFWTechUpdate}
G.~B. H.~E. Learned-Miller.
\newblock Labeled faces in the wild: Updates and new reporting procedures.
\newblock Technical Report UM-CS-2014-003, University of Massachusetts,
  Amherst, May 2014.

\bibitem{Lezama16}
J.~Lezama, Q.~Qiu, and G.~Sapiro.
\newblock Not afraid of the dark: {NIR-VIS} face recognition via cross-spectral
  hallucination and low-rank embedding.
\newblock In {\em CVPR}, 2017.

\bibitem{Li2008}
J.~Li, P.~Hao, C.~Zhang, and M.~Dou.
\newblock Hallucinating faces from thermal infrared images.
\newblock In {\em ICIP}, pages 465--468, 2008.

\bibitem{RedNet16}
X.-J. {Mao}, C.~{Shen}, and Y.-B. {Yang}.
\newblock {Image Restoration Using Convolutional Auto-encoders with Symmetric
  Skip Connections}.
\newblock In {\em NIPS}, 2016.

\bibitem{Mirza2014}
M.~Mirza and S.~Osindero.
\newblock Conditional generative adversarial nets.
\newblock In {\em ArXiv}, 2014.

\bibitem{Mostafa13}
E.~Mostafa, R.~Hammoud, A.~Ali, and A.~Farag.
\newblock Face recognition in low resolution thermal images.
\newblock {\em Computer Vision and Image Understanding}, 117(12):1689 -- 1694,
  2013.

\bibitem{irisdataset}
U.~of~Tennessee.
\newblock {\em EEE OTCBVS WS Series Bench; DOE University Research Program in
  Robotics under grant DOE-DE-FG02-86NE37968}.
\newblock 2012.

\bibitem{Parkhi15}
O.~M. Parkhi, A.~Vedaldi, and A.~Zisserman.
\newblock Deep face recognition.
\newblock In {\em British Machine Vision Conference}, 2015.

\bibitem{Radford2016}
A.~Radford, L.~Metz, and S.~Chintala.
\newblock Unsupervised representation learning with deep convolutional
  generative adversarial networks.
\newblock In {\em ICLR}, 2016.

\bibitem{Reale2014}
C.~Reale, N.~M. Nasrabadi, and R.~Chellappa.
\newblock Coupled dictionaries for thermal to visible face recognition.
\newblock In {\em ICIP}, pages 328--332, 2014.

\bibitem{Riggan2016}
B.~S. Riggan, N.~J. Short, S.~Hu, and H.~Kwon.
\newblock Estimation of visible spectrum faces from polarimetric thermal faces.
\newblock In {\em BTAS}, pages 1--7, 2016.

\bibitem{Ronneberger2015}
O.~Ronneberger, P.~Fischer, and T.~Brox.
\newblock U-net: Convolutional networks for biomedical image segmentation.
\newblock In {\em MICCAI}, 2015.

\bibitem{Sarfraz2017}
M.~S. Sarfraz and R.~Stiefelhagen.
\newblock Deep perceptual mapping for cross-modal face recognition.
\newblock {\em IJCV}, 122(3):426--438, 2017.

\bibitem{Schroff15}
F.~Schroff, D.~Kalenichenko, and J.~Philbin.
\newblock Facenet: A unified embedding for face recognition and clustering.
\newblock In {\em CVPR}, pages 815--823, 2015.

\bibitem{Nathaniel15}
N.~Short, S.~Hu, P.~Gurram, K.~Gurton, and A.~Chan.
\newblock Improving cross-modal face recognition using polarimetric imaging.
\newblock {\em Opt. Lett.}, 40(6):882--885, 2015.

\bibitem{Simonyan2015}
K.~Simonyan and A.~Zisserman.
\newblock Very deep convolutional networks for large-scale image recognition.
\newblock In {\em ICLR}, 2015.

\bibitem{Tran2017}
L.~Tran, X.~Yin, and X.~Liu.
\newblock Disentangled representation learning gan for pose-invariant face
  recognition.
\newblock In {\em CVPR}, 2017.

\bibitem{Wang2016}
X.~Wang and A.~Gupta.
\newblock Generative image modeling using style and structure adversarial
  networks.
\newblock In {\em ECCV}, 2016.

\bibitem{Wen2016}
Y.~Wen, K.~Zhang, Z.~Li, and Y.~Qiao.
\newblock A discriminative feature learning approach for deep face recognition.
\newblock In {\em ECCV}, pages 499--515, 2016.

\bibitem{2014Yi}
D.~{Yi}, Z.~{Lei}, S.~{Liao}, and S.~Z. {Li}.
\newblock {Learning Face Representation from Scratch}.
\newblock {\em ArXiv e-prints}, 2014.

\bibitem{Yi2007}
D.~Yi, R.~Liu, R.~Chu, Z.~Lei, and S.~Z. Li.
\newblock Face matching between near infrared and visible light images.
\newblock In {\em ICB}, pages 523--530, 2007.

\bibitem{ZhangIJCB2017}
H.~{Zhang}, V.~M. {Patel}, B.~S. {Riggan}, and S.~{Hu}.
\newblock Generative adversarial network-based synthesis of visible faces from
  polarimetric thermal faces.
\newblock In {\em IJCB}, 2017.

\bibitem{Zhang2017}
H.~Zhang, T.~Xu, H.~Li, S.~Zhang, X.~Wang, X.~Huang, and D.~Metaxas.
\newblock Stackgan: Text to photo-realistic image synthesis with stacked
  generative adversarial networks.
\newblock In {\em ICCV}, 2017.

\end{thebibliography}
